\documentclass[conference]{IEEEtran}
\usepackage{cite}
\usepackage{amsmath,amssymb,amsfonts}
\usepackage{algorithmic}
\usepackage{graphicx}
\usepackage{textcomp}
\usepackage{xcolor}
\usepackage{subfig}
\usepackage{multicol}

\graphicspath{ {images/} }

\AtBeginDocument{%
  }
\begin{document}
\nocite{*}

\title{Homography Estimation in Complex Topological Scenes\\}

\author{\IEEEauthorblockN{1\textsuperscript{st} Giacomo D'Amicantonio}
\IEEEauthorblockA{\textit{Eindhoven University of Technology} \\
Eindhoven, Neatherlands \\
g.d.amicantonio@tue.nl}
\and
\IEEEauthorblockN{2\textsuperscript{nd} Egor Bondarau}
\IEEEauthorblockA{\textit{Eindhoven University of Technology} \\
Eindhoven, Neatherlands \\
e.bondarau@tue.nl}
\and
\IEEEauthorblockN{3\textsuperscript{rd} Peter H.N. De With}
\IEEEauthorblockA{\textit{Eindhoven University of Technology} \\
Eindhoven, Netherlands \\
p.h.n.de.with@tue.nl}
}

\maketitle

\begin{abstract}
Surveillance videos and images are used for a broad set of applications, ranging from traffic analysis to crime detection. Extrinsic camera calibration data is important for most analysis applications. However, security cameras are susceptible to environmental conditions and small camera movements, resulting in a need for an automated re-calibration method that can account for these varying  conditions.
In this paper, we present an automated camera-calibration process leveraging a dictionary-based approach that does not require prior knowledge on any camera settings. The method consists of a custom implementation of a Spatial Transformer Network (STN) and a novel topological loss function. Experiments reveal that the proposed method improves the IoU metric by up to 12\% w.r.t. a state-of-the-art model across five synthetic datasets and the World Cup 2014 dataset. 
\end{abstract}


\section{Introduction}
The topic of camera calibration has been of great interest in the Computer Vision community for decades, being often the first building block of an application. Calibration data are required for applications such as sports video broadcasting, object localization and immersive imaging. A multitude of methods and algorithms have been developed to perform automated calibration in different contexts. Unfortunately, these methods are infeasible or impractical in real-world setups. For example, the conventional checkerboard methods work well in lab conditions, but are inapplicable for cameras on roads or traffic intersections. For autonomous driving, we need real-time data on stationary topology and temporal objects positioning at intersections. Analysis of feeds from cameras located at 5-7 meters height can provide both of these types of data in real-time. However, an accurate calibration of these cameras is a prerequisite. Automated and frequent re-calibration brings robustness and accuracy in this approach, since cameras a always moving under different forces (e.g. wind, close-passing trucks). Automated camera calibration remains an open problem and there are no proven robust methods applied in the traffic surveillance industry.

In the past decade, several deep learning approaches have been developed and applied to camera calibration. The main problem of these methods is that they rely on large datasets of annotated data, which is a solid constraint for small companies and researchers. This paper proposes a homography estimation method relying on synthetic data without the need for the ground-truth homography. 
The proposed method contributes in two aspects:

\begin{itemize}
    \item novel loss function, called Topological Loss, to train a Spatial Transformer Network (STN) model exploiting the structure of the scene;
    \item  improvement of the STN architecture, specifically of the localization layers of the network.
\end{itemize}
We train and test our model on synthetic datasets created on the bird's-eye view images of five road intersections and on the World Cup 2014 dataset presented in \cite{Homayounfar_2017_CVPR}. We show that our homography estimation performs well in complex contexts such as road intersections. 

The paper is structured as follows. Section~2 presents existing methods related to automated camera calibration and homography estimation. Section~3 describes the dataset generation methodology, the homography estimation model and the loss function. Section~4 explains the experimental settings for different models, loss functions and training strategies. Section~5 discusses the ablation study and the results. Section~6 concludes the paper. 

\section{Related Work}
Two classes of homography estimation techniques exist: conventional and deep learning-based methods. Among the conventional methods, the work in~\cite{6163012} uses large calibration patterns from roads, buildings and other landmarks, instead of checkerboard patterns placed in front of a camera.
In~\cite{42}, the authors propose three vanishing points to calibrate the camera estimated with a KLT tracker (\cite{323794}, \cite{Tomasi91detectionand}, \cite{LBKT}) and the Hough transform. 
In~\cite{07243} and~\cite{015}, the approach estimates vanishing points based on car trajectories to obtain the extrinsic parameters. For the intrinsic parameters, the authors rely on conventional methods and assumptions, e.g. compute only focal length and fish-eye distortion, respectively. The work in~\cite{3199667} exploits a combination of Faster R-CNN~\cite{RenHGS15} and background subtraction to extract selected key-point features from vehicles in the video and estimate the calibration parameters using an aggregating and filtering algorithm.
Deep learning methodologies have proven to be more successful for camera calibration. 
In~\cite{PTZ}, the authors exploit two manually annotated 3D-2D matching points and estimate pan/tilt camera angles with one of the two points, using the other point to refine the prediction, with a consecutive random forest algorithm to complete the calibration. \\
\begin{figure}
\centering
    \subfloat{\includegraphics[width=0.235\textwidth]{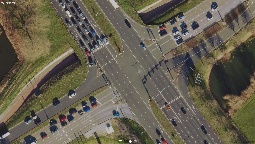}}\hfill
    \subfloat{\includegraphics[width=0.235\textwidth]{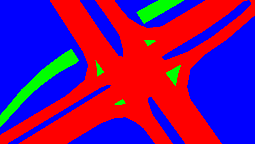}}\hfill
    \subfloat{\includegraphics[width=0.235\textwidth]{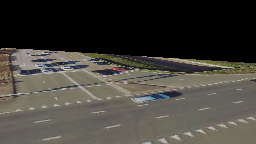}}\hfill
    \subfloat{\includegraphics[width=0.235\textwidth]{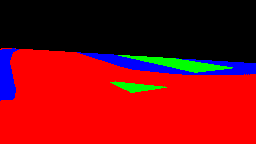}}\hfill
    \caption{Bird's-eye view and warped view of an intersection.
    Left column: bird's-eye view of an intersection and its semantically segmented map. Right column: obtained images after warping the bird's-eye view and its map with a sampled homography. The images are used to train the homography estimation model.}
    \label{fig:warping}
\end{figure}

\begin{figure}[hb]
\centering
    \subfloat{\includegraphics[width=0.235\textwidth]{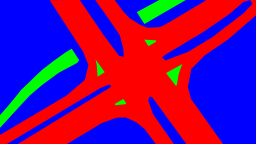}}\hfill
    \subfloat{\includegraphics[width=0.235\textwidth]{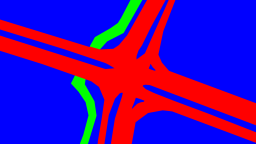}}\hfill
    \subfloat{\includegraphics[width=0.235\textwidth]{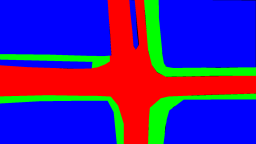}}\hfill
    \subfloat{\includegraphics[width=0.235\textwidth]{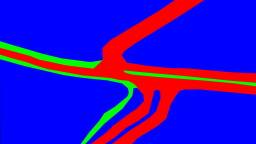}}\hfill
    \subfloat{\includegraphics[width=0.235\textwidth]{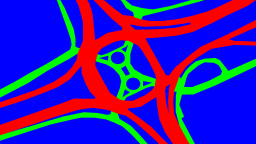}}\hfill
\caption{Bird's-eye views of the intersections, semantically segmented according to three semantic regions: (a) road, (b) terrain and (c) bicycle path. The intersection are numbered 1 through 5 from top to bottom and left to right. Notice that the first two intersections have similar topologies, while the last has a much more complex topology than the others.} 
\end{figure}
\indent
In~\cite{9025456}, the authors propose to calibrate sports cameras on synthetic data. The model matches the input image with a set of templates and refines the pose of the input starting from the matched template. The proposed network learns to extract the edges of the field markings on a football stadium and encodes them to a feature representation, which is matched with a database of known synthetic images. Finally, the authors refine the estimated pose using the matched image. 
In~\cite{9157432}, the authors deploy a similar framework to estimate the homography for broadcast cameras in football and basketball fields. Instead of using the edge image, they feed the semantically segmented image to a Siamese network that matches it with a dictionary of images having known homographies. The model concatenates the semantic maps with its matched template and then a Spatial Transformer Network (STN)~\cite{STN-Homography} estimates the homography between the semantic map and its matched template. A similar approach can be found in~\cite{2969465}.\\
\indent
In this paper, we extend the homography estimation approach on traffic intersections. This choice is motivated by the fact that it allows to calibrate the camera based on the geographical coordinates of the camera location only. Furthermore, moving actors in the scene and weather conditions can influence the camera calibration process and cannot be controlled. Semantically segmented images can help in reducing the influence of these factors. 

\section{Method}
\begin{figure*}
\begin{center}
\includegraphics[width=1\textwidth]{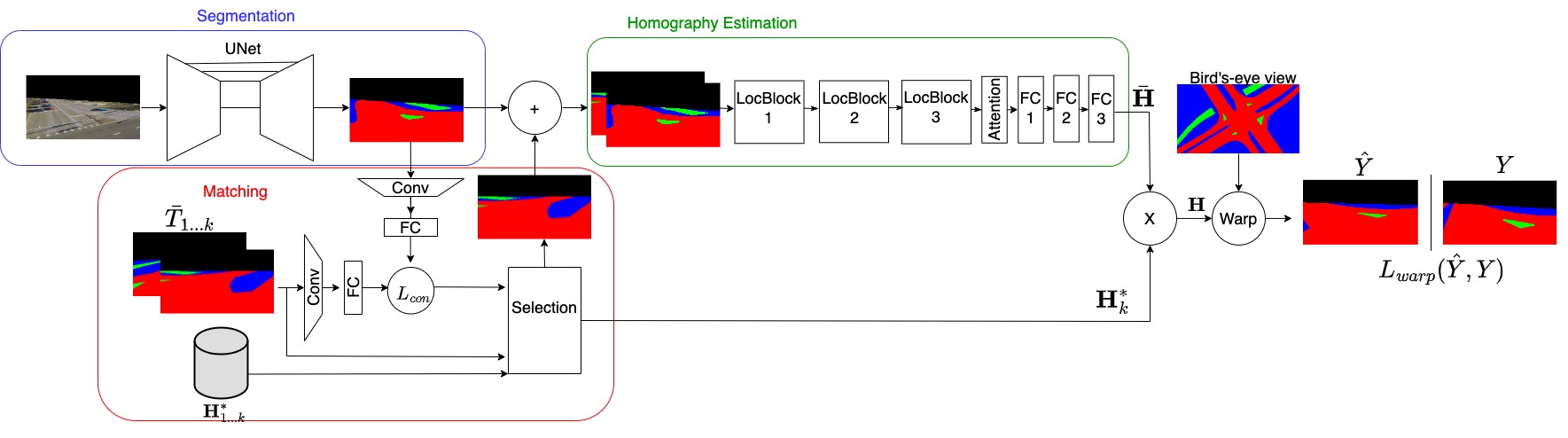}
\end{center}
   \caption{Architecture of the proposed model. The input images from the camera to be calibrated are semantically segmented by UNet to create the semantically segmented image~$\bar{Y}$. The Siamese network finds the closest match between~$\bar{Y}$ and the dictionary of templates~$T_{1...k}$. The two images are then concatenated across the channel dimension and supplied to the STN. The homography $\hat{\mathbf{H}}$ is estimated by three so-called Localization Blocks (LocBlock) and three fully connected (FC) layers. The two matrices~$\hat{\mathbf{H}}$ and~$\mathbf{H}_K^*$ are multiplied to produce the final homography~$\mathbf{H}$. The model warps the bird's-eye view with~$\mathbf{H}$ to generate the image~$\hat{Y}$.}
\label{fig:structure}
\end{figure*}

The overall architecture of the proposed model is presented in Figure~\ref{fig:structure}. Due to the lack of publicly available datasets, the model is trained and tested on the synthetic datasets generated as described in Section~3.1. The pipeline is composed of Semantic Segmentation, Matching, Homography Estimation and Warping. 
\subsection{Data Generation}
First, a synthetic dataset is generated for each of the intersections and the soccer field for experimenting. The bird's-eye view image of the scene and its semantically segmented map are warped with the homographies created with parameter settings explained later. Examples of the resulting images are shown in Figure~\ref{fig:warping}. The dataset contains the input images, the semantic ground-truth data and the homographies used to warp them. We consider the virtual camera as a PTZ camera model with a projective matrix $\mathbf{P}$, which is defined by: 
\begin{equation}
\mathbf{P} = \mathbf{K}\cdot\mathbf{R}\cdot\left[\mathbf{I}\,|\mathbf{C}\right] = \mathbf{K}\cdot\mathbf{Q}\cdot\mathbf{S}\cdot\left[\mathbf{I}\,|\mathbf{C}\right].
\end{equation}
In the equation, $\mathbf{K}$ denotes the intrinsic matrix, $\mathbf{I}$ is the identity matrix and $\mathbf{C}$ is the camera translation matrix. The rotation matrix $\mathbf{R}$ is decomposed into $\mathbf{Q}$, the rotation of the camera due to pan and tilt of the camera orientation, and $\mathbf{S}$, being the rotation of the camera w.r.t. the world coordinates. The virtual camera is rotated by -90\textdegree along the $y$-axis of the world surface, so that the camera projection is parallel to the ground plane.

\begin{table}[hb]
\caption{Sampling grid for the soccer dataset}
\centering
\begin{tabular}{|l|c|c|}
    \hline 
    \textbf{Parameters} & \textbf{Range} & \textbf{Step size}\\
    \hline
    Pan & (-25\textdegree, +25\textdegree) & 1\textdegree \\
    Tilt & (-15\textdegree, 0\textdegree) & 1\textdegree \\
    Focal length & (500, 800) & 50\\
    \hline
\end{tabular}
\label{tab:samplingsb}
\end{table}

\begin{table}[hb]
\caption{Sampling grid for the intersections datasets.}
\centering
\begin{tabular}{|l|c|c|}
    \hline
    \textbf{Parameters} & \textbf{Range} & \textbf{Step size}\\
    \hline
    Pan & (-180\textdegree, +180\textdegree) & 15\textdegree \\
    \hline
    Tilt & (-20\textdegree, 0\textdegree) & 5\textdegree \\
    \hline
    Focal length & (50, 500) & 50\\
    \hline
    X & (600, 700) & 10 \\
    \hline
    Y & (900, 1000) & 10 \\
    \hline
    Z & (50, 100) & 10 \\
    \hline
\end{tabular}
\label{tab:samplingint}
\end{table}

\indent
To create the synthetic datasets, we define a sampling grid composed of focal length $f$, tilt and pan angles, as described in~\cite{9157432}. For the World Cup 2014 dataset, we only need to create a synthetic dictionary. To do so, we align the camera location to one of the two long sides of the field and generate 4,500~homographies by sampling the grid with parameters as listed in Table~\ref{tab:samplingsb}.
The procedure to generate the synthetic dataset for intersections is similar. In addition to the above-explained three parameters, we sample the X, Y and Z camera coordinates. The virtual cameras are considered to be located in the central areas of the intersections. We generate 200k~homographies by sampling from the grid with parameters specified in Table~\ref{tab:samplingint}.\\
\indent
For the intersections, we generate and we randomly select 3,000, 500 and 1,000 images for the training set, test set and dictionary, respectively. This is done for two main reasons. First, a larger dictionary means that the matching process by the Siamese network takes considerably more time, given that every input image has to be compared with each image in the dataset. Second, we want also to evaluate the performance of the proposed model when the matching is not guaranteed to be optimal. This often occurs in our case with intersection datasets, where the dictionary contains only 0.5\% of the total homographies that are generated.

\subsection{Topological Loss}
\begin{figure}[hb]
\centering
    \includegraphics[width=0.4\textwidth]{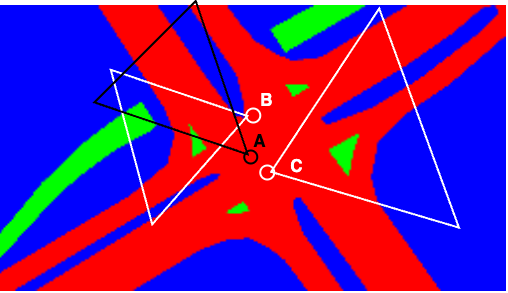}
    \caption{Positions and view of the three virtual cameras in the bird-eye-view image.}  \label{fig:similarityc}
\end{figure}

\begin{figure}[t]
\centering
    \includegraphics[width=0.45\textwidth]{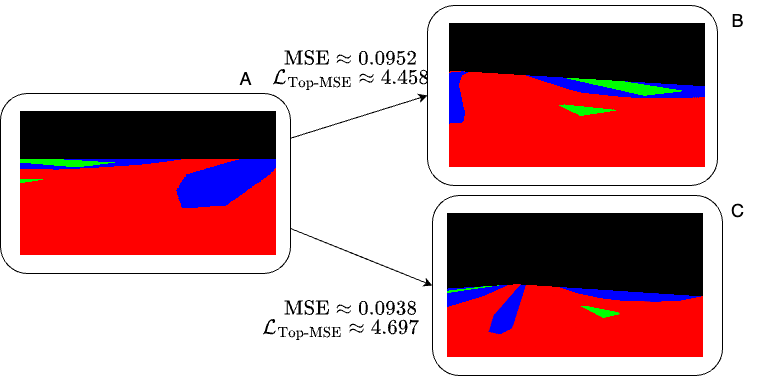}
    \caption{$\text{MSE}$ and $\mathcal{L}_{\text{Top-MSE}}$ scores between different images of an intersection obtained by sampling homographies.}
    \label{fig:similaritya}
\end{figure}

If the topology of a scene is such that two cameras are oriented in two different directions and yet capture two similar images, as in Figure~\ref{fig:similarityc}, the resulting pixel-based error can be very small. Pixel-by-pixel loss functions can fail to address the structural similarity between images, as shown in Figure~\ref{fig:similaritya}.  In such a case, the STN would learn that the estimated homography is very close to the correct one while in reality it is very different. \\
\indent
To address this problem, we introduce a new loss function to train the STN and generate the ground truths for the matching component, called Topological Loss ($\mathcal{L}_{\text{Top}}$). The Topological Loss splits the predicted $\hat{Y}$ and the ground truth $Y$ in $N$ patches $\hat{Y}_{i,j}$ and $Y_{i,j}$ while maintaining the original aspect ratio. The loss between corresponding patches is computed by any pixel-based loss function, \emph{e.g.} the Mean-Squared Error (MSE). The Topological loss, implemented with MSE, is denoted as $\mathcal{L}_{\text{Top-MSE}}$ and is made by accumulating patch losses. Similarly, $\mathcal{L}_{\text{Top-Dice}}$ is the Topological loss implemented with Dice loss.
The loss of every patch is increased if the losses of the neighbouring patches are higher than a threshold $\beta$, which is defined by:
\begin{align}
\label{eqn:TLij}
    \mathcal{L}_{\text{patch}}(\hat{Y}, Y) &= \text{MSE}(\hat{Y}_{i,j}, Y_{i,j}) + \\ \nonumber
    & \alpha\sum_{k}\sum_{l}\max{(0, \text{MSE}(\hat{Y}_{i+k,j+l}, Y_{i+k,j+l})-\beta}),
\end{align}
where the running indexes $k, l\in[-1, 0, 1]$. The final loss $\mathcal{L}_{\text{Top-MSE}}$ becomes now as follows:
\begin{equation}
    \mathcal{L}_{\text{Top-MSE}} = \frac{1}{N}\sum_{i}\sum_{j}\mathcal{L}_{\text{patch}}(\hat{Y}_{i,j}, Y_{i,j}),
\end{equation}
where $i, j\in[0, ..., \sqrt{N}]$.

\begin{figure}[t]
\centering
    \includegraphics[width=0.45\textwidth]{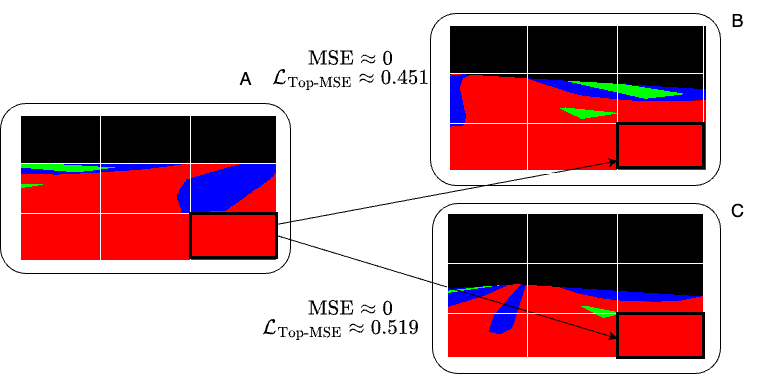}
    \caption{$\text{MSE}$ and $\mathcal{L}_{\text{Top-MSE}}$ scores between two patches. Notice that, using $\text{MSE}$, the patch would be considered almost completely correct while using $\mathcal{L}_{\text{Top-MSE}}$, the error is quite large.}  \label{fig:similarityb}
\end{figure}

As shown in Figure~\ref{fig:similarityb}, segmented images may contain many pixels of the same semantic region type that actually correspond to different real-world regions. Still, those points would lead to a low MSE score. With $\mathcal{L}_{\text{Top-MSE}}$, those pixels generate a higher loss due to the difference between the neighbouring patches. The parameter $\alpha$ in Equation~(\ref{eqn:TLij}) allows to balance the influence of neighbouring patches over the loss function. 
\subsection{Segmentation}
We identify three semantic areas in each context and the additional background area, making each semantic map a 4-channel image. Actors that can interfere with the calibration, like cars in an intersection or players on the soccer field, are removed by the segmentation model. The segmentation step allows the subsequent components of the model to focus on the semantically segmented structure of the scene. \\
\indent
To perform the semantic segmentation, we deploy a UNet~\cite{10.1007/978-3-319-24574-4_28} encoder-decoder model. An image~$I$ is given as input and the model produces a semantic map~$\bar{Y}$. The UNet is trained with the standard cross-entropy loss between the predicted~$\bar{Y}$ and the semantic ground-truth map~$Y$.
\subsection{Matching}
As mentioned in Section~3.1, the generated dataset is split into train set, test set and dictionary. Each entry in the dictionary is composed of pairs of a semantic template~$T_k$ and its homography~$\mathbf{H}_k^*$. The Siamese network finds the closest match between the semantic map~$\bar{Y}$ and the templates in the dictionary.  The network is trained to match the images using the Contrastive loss~\cite{contrastive}.
The correct match for an input image $I$ is the the template-homography pair $(\bar{T_k}, \mathbf{H}_k^*)$ that has the lowest pixel-based distance to $I$, calculated with a matching function such as the Mean Squared Error (MSE). 

\subsection{Homography Estimation}
\begin{figure}[t]
   \includegraphics[scale=0.16]{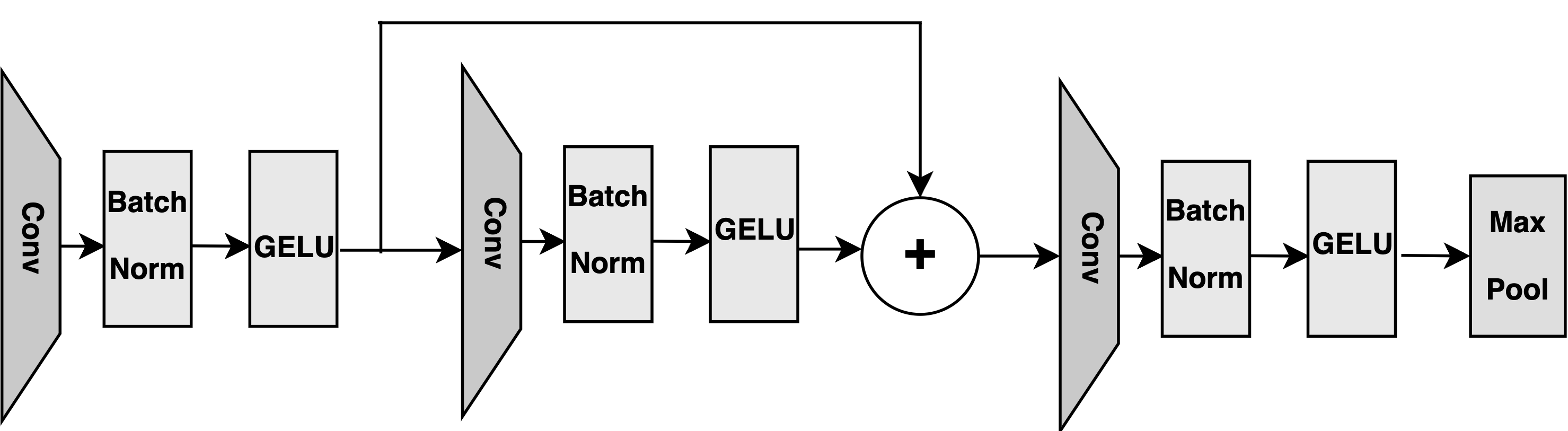}
   \caption{Architecture of a Localization Block (LocBlock). The residual connection allows the LocBlock to learn structural similarities between feature maps at different scales.} 
   \label{fig:Locblock}
\end{figure}

The semantic map $\bar{Y}$ produced by the UNet is concatenated across the channels dimension with the template $\bar{T}_k$, which was previously matched by the Siamese network, creating an 8-channel image. The image is supplied to the Spatial Transformer Network (STN)~\cite{2969465}.
The first part of the STN usually consists of two convolutional layers, called location layers. This first part is replaced by three  novel parts, called Location Blocks (LocBlock), as shown in Figure~\ref{fig:Locblock}. The major differences between Location Blocks and location layers are the residual connections, which help the network to learn similarities between feature maps at different scales in images with complex topologies. After the localization blocks, a self-attention mechanism \cite{NIPS2017_3f5ee243} is added to learn global dependencies across the feature map produced.  \\
\indent
The last fully connected layer estimates the first 8~values of a 3$\times$3 homography $\mathbf{\bar{H}}$, which is the relative transformation between $\bar{Y}$ and the template $\hat{T}_k$. The final homography becomes $\mathbf{H} = \hat{\mathbf{H}}_k\bar{\mathbf{H}}$. We initialize the weights of the last fully connected layer to zero and the biases to unity in accordance with the elements at the main diagonal of an identity matrix. This is an optimal initialization for the STN because the Siamese network finds the optimal match $\bar{T}_k$ for $\bar{Y}$ and its homography $\bar{\mathbf{H}}_k$. Therefore, the first homography that the network employs to warp the bird's-eye view is the template homography $\mathbf{H}=\bar{\mathbf{H}}_k\mathbf{I}=\bar{\mathbf{H}}_k$. During training, the STN learns to modify $\bar{\mathbf{H}}_k$ from the matched template to obtain a better estimation of the homography of the input image.

\begin{figure}[b]
\includegraphics[scale=0.28]{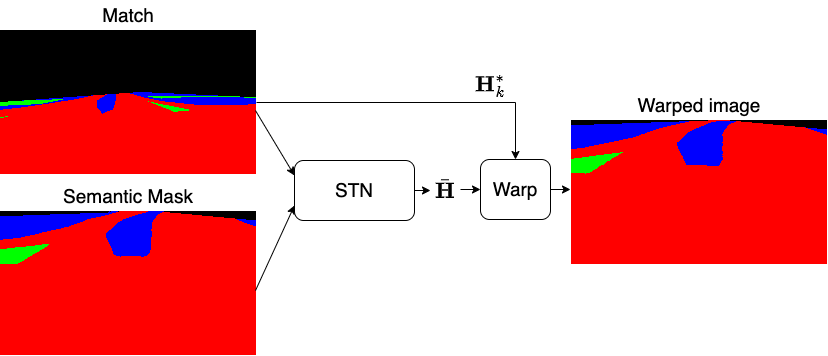}
   \caption{Visual example of the homography estimation on the 2nd intersection dataset, showing multiple instances of the same semantic region, starting from a suboptimal match.}
\label{fig:warpexample}
\end{figure}

\subsection{Training}
Each component of the network can be trained with its own loss function (step-by-step), or all the components can be trained jointly (end-to-end). End-to-end training means that the components are trained at the same time with a joint loss function composed of loss functions for each component. In addition to the earlier mentioned loss component for the UNet and Siamese networks, the STN can be trained with a generic loss term based on image warping, called  $\mathcal{L}_{warp}$. In such a setting, the overall loss function of the model becomes:
\begin{equation}
    \mathcal{L} = \eta \cdot \mathcal{L}_{CE} + \tau \cdot \mathcal{L}_{con} + \gamma \cdot \mathcal{L}_{warp} \ ,
    \label{eqn:TL}
\end{equation}
where the subscript CE stands for cross-entropy. The weight factors $\eta, \tau$ and $\gamma$ can be adjusted to balance the influence of each individual loss and will be specified in the next section.

\section{Experimental Setup and Parameters}
We experiment on the synthetic datasets generated for five intersections, the World Cup 2014 dataset and three mixtures of the synthetic datasets. It is worth noting that using synthetic datasets does not prevent us from evaluating the performances of the proposed model and the Topological loss for two reasons. Firstly, the segmentation component of the proposed model is identical to the segmentation component of the baseline model, which allows for a fair comparison between the improvements brought by our proposals. Furthermore, the segmentation component can be updated with more recent networks and fine-tuned.
The proposed model is trained with the conventional versions of MSE and Dice Loss as well as their Topological Loss implementations, $\mathcal{L}_{\text{Top-MSE}}$ and $\mathcal{L}_{\text{Top-Dice}}$. In all experiments, the coefficient $\alpha$ and the threshold $\beta$ in Equation~(\ref{eqn:TLij}) are both set to $\alpha,\beta$=0.3 and the number of patches~$N$ in Equation~(\ref{eqn:TL}) is set to $N$=16. To evaluate the performances of Topological Loss for the matching component, we experiment by using MSE and $\mathcal{L}_{\text{Top-MSE}}$ to create the ground truths. \\

\begin{table}[t]
\caption{Warmup schedule and parameters setting for end-to-end training of the complete model.}
\vspace*{0.1cm}
\centering
\begin{tabular}{|l|c|c|c|c|}
    \hline
    \textbf{Model} & \textbf{Epochs} & $\eta$ & $\tau$ & $\gamma$\\
    \hline
    UNet & $\leq$ 20 & 1 & 0 & 0 \\
    \hline
    Siamese & 20-40 & 0 & 1 & 0 \\
    \hline
    End-to-end & $\ge$ 40 & 0.05 & 0.05 & 0.9 \\
    \hline
\end{tabular}
\label{tab:trainingsched}
\end{table}

We also experiment and compare with the model proposed by Sha et~al.~\cite{9157432}, which we consider the baseline model.
In every experiment, we warm up the UNet and the Siamese network for 20~epochs each. This scheduling is necessary because if the semantic images produced by the UNet are of a very low quality at the early stages of the training, the Siamese network learns incorrect features. This means that the templates selected by the Siamese network become incorrect for the STN and the network starts to learn wrong features. Finally, we train every combination of the proposed model and the baseline model with the four forms of the loss function in step-by-step and end-to-end fashion. \\
\indent
We evaluate the performances with the Intersection-over-Union~(IoU) metric. IoU shows the similarity between the images produced by warping the bird's-eye view with the estimated homographies from the model and the ground-truth used to train the UNet. \\
\indent
All models are implemented in PyTorch and Kornia. The synthetic datasets are randomly split in train set, test set and dictionary, composed of~3000, 500 and 1000~samples, respectively. All the experiments are executed on a computer equipped with a Titan~RTX GPU and 32~GB of memory. We report the mean performances obtained for each experimental setup involving a specific combination, for five complete cycles of dataset sampling, training and measuring results. 

\section{Experiments and Ablation Study}
This section presents results of the experiments on the datasets. The perfomance of Topological Loss is separately evaluated on the matching and the homography estimation tasks. The proposed model is evaluated in comparison to the previous SOTA model for homography estimation in the context of scene topology presented by Sha et al. in \cite{9157432}. The experiments are performed on the World Cup 2014 dataset, the five synthetic intersection datasets described in Section 3.1 and on three mixes of these dictionaries.

\begin{table*}[h]
\caption{Comparison of IoU scores for the World Cup 2014 and the intersections datasets with different matching functions.}
\centering
\begin{tabular}{|l|l|l|c|c|c|c|}
    \hline
    \textbf{Dataset} & \textbf{Matching} & \textbf{Method} & \multicolumn{4}{|c|}{\textbf{Measured IoU}}\\
    \cline{4-7}
    & \textbf{Score} & & $\mathcal{L}_{\text{Top-MSE}}$ & MSE & $\mathcal{L}_{\text{Top-Dice}}$ & Dice\\
    \hline
    & & Ours$_{EtE}$ & \textbf{83.01\%} & 82.73\% & 82.98\% & 81.98\% \\
    \cline{3-7}
    World Cup 2014 & MSE & Ours & 82.81\% & 82.77\% & \textbf{83.06\%} & 79.23\% \\
    \cline{2-7}
     &  & Ours$_{EtE}$ & \textbf{84.21\%} & 83.41\% & 83.76\% & 82.91\% \\
    \cline{3-7}
     & $\mathcal{L}_{\text{Top-MSE}}$ & Ours & \textbf{83.92\%} & 83.29\% & 83.28\% & 82.55\% \\
    \hline
    & & Ours$_{EtE}$ & \textbf{87.11\%} & 84.32\% & 85.49\% & 84.73\% \\
    \cline{3-7}
    Intersection N.1 & MSE & Ours & \textbf{86.67\%} & 85.36\% & 86.52\% & 84.98\%\\
    \cline{2-7}
     &  & Ours$_{EtE}$ & 87.42\% & 84.28\% & \textbf{87.57\%} & 85.53\%\\
    \cline{3-7}
     & $\mathcal{L}_{\text{Top-MSE}}$ & Ours & \textbf{87.91\%} & 85.65\% & 87.44\% & 86.77\%\\
    \hline
    & & Ours$_{EtE}$ & 84.03\% & 83.07\% & \textbf{85.35\%} & 83.48\% \\
    \cline{3-7}
    Intersection N.2 & MSE  & Ours & 84.43\% & 82.71\% & \textbf{84.57\%} & 83.04\%\\
    \cline{2-7}
     &  & Ours$_{EtE}$ & 85.12\% & 83.29\% & \textbf{87.00\%} & 84.71\% \\
    \cline{3-7}
     & $\mathcal{L}_{\text{Top-MSE}}$ & Ours & 85.22\% & 83.82\% & \textbf{85.48\%} & 84.23\% \\
    \hline
    & & Ours$_{EtE}$ & \textbf{84.74\%} & 74.25\% & 84.25\% & 80.37\%\\
    \cline{3-7}
    Intersection N.3 & MSE & Ours & 82.04\% & 80.48\% & \textbf{82.58\%} & 79.36\%\\
    \cline{2-7}
     & & Ours$_{EtE}$ & 85.42\% & 77.33\% & \textbf{85.93\%} & 84.51\% \\
    \cline{3-7}
     & $\mathcal{L}_{\text{Top-MSE}}$ & Ours & 82.79\% & 78.98\% & \textbf{86.66\%} & 79.48\% \\
    \hline
    &  & Ours$_{EtE}$ & 81.38\% & 80.25\% & \textbf{82.24\%} & 80.57\%\\
    \cline{3-7}
    Intersection N.4 & MSE & Ours & 81.57\% & 80.17\% & \textbf{83.05\%} & 73.91\%\\
    \cline{2-7}
     & & Ours$_{EtE}$ & 83.61\% & 82.64\% & \textbf{83.72\%} & 82.52\%\\
    \cline{3-7}
     & $\mathcal{L}_{\text{Top-MSE}}$  & Ours & 82.22\% & 78.98\% & \textbf{84.51\%} & 82.23\%\\
    \hline
    & & Ours$_{EtE}$ & 68.57\% & 58.45\% & \textbf{73.13\%} & 63.71\%\\
    \cline{3-7}
    Intersection N.5 & MSE & Ours & 66.19\% & 59.57\% & \textbf{72.04\%} & 60.99\%\\
    \cline{2-7}
     & & Ours$_{EtE}$ & 69.71\% & 60.55\% & \textbf{73.13\%} & 63.71\% \\
    \cline{3-7}
     & $\mathcal{L}_{\text{Top-MSE}}$ & Ours & 71.11\% & 60.03\% & \textbf{73.75\%} & 62.76\% \\
    \hline
\end{tabular}
\label{tab:largematchingcompare}
\end{table*}

\subsection{Matching with Topological Loss}
In all setups, the use of Topological Loss shows improvements compared to the same model trained with the respective conventional version of the loss function, as shown in Table~\ref{tab:largematchingcompare}. It can be observed that the models trained with $\mathcal{L}_{\text{Top-MSE}}$ and $\mathcal{L}_{\text{Top-Dice}}$ are the top-2 performing networks in every combination and that using $\mathcal{L}_{\text{Top-MSE}}$ to generate the ground truth for the Siamese network results in an improvement of 1\% on average. Given the consistency of the perfomance improvements obtained by matching with $\mathcal{L}_{\text{Top-MSE}}$, we will use this version of the model in further comparison experiments. 

\begin{table*}[h]
\caption{IoU scores obtained by the proposed model and the baseline model.}
\centering
\begin{tabular}{|l|l|c|c|c|c|}
    \hline
    {\textbf{Dataset}} & {\textbf{Method}} & \multicolumn{4}{|c|}{\textbf{Measured IoU}}\\
    \cline{3-6}
     & & $\mathcal{L}_{\text{Top-MSE}}$ & MSE & $\mathcal{L}_{\text{Top-Dice}}$ & Dice\\
    \hline
    & Sha et al.$_{EtE}$ & \textbf{81.67\%} & 80.12\% & 80.65\% & 79.89\% \\
    \cline{2-6}
     World Cup 2014 & Sha et al. & \textbf{81.64\%} & 80.6\% & 81.55\% & 80.09\%  \\
    \cline{2-6}
     & Ours$_{EtE}$ & \textbf{84.21\%} & 83.41\% & 83.76\% & 82.91\%\\
    \cline{2-6}
     & Ours & \textbf{83.92\%} & 83.29\% & 83.28\% & 82.55\%\\
    \hline
     & Sha et al.$_{EtE}$ & \textbf{84.62\%} & 83.59\% & 83.25\% & 82.03\% \\
    \cline{2-6}
    Intersection N.1 & Sha et al. & 85.94\% & 84.41\% & \textbf{85.96\%} & 84.72\% \\
    \cline{2-6}
     & Ours$_{EtE}$ & 87.42\% & 84.28\% & \textbf{87.57\%} & 85.53\% \\
    \cline{2-6}
     & Ours & \textbf{87.91\%} & 85.65\% & 87.44\% & 86.77\%\\
    \hline
     & Sha et al.$_{EtE}$ & 75.93\% & 75.15\% & \textbf{76.18\%} & 74.77\% \\
    \cline{2-6}
    Intersection N.2 & Sha et al. & 78.04\% & 78.46\% & \textbf{78.42\%} & 77.55\% \\
    \cline{2-6}
     & Ours$_{EtE}$ & 85.12\% & 83.29\% & \textbf{87.00\%} & 84.71\% \\
    \cline{2-6}
     & Ours & 85.22\% & 83.82\% & \textbf{85.48\%} & 84.23\% \\
    \hline
     & Sha et al.$_{EtE}$ & 76.22\% & 75.42\% & \textbf{77.49\%} & 76.29\% \\
    \cline{2-6}
    Intersection N.3 & Sha et al. & 78.98\% & 78.53\% & \textbf{78.99\%} & 78.73\% \\
    \cline{2-6}
     & Ours$_{EtE}$ & 85.42\% & 77.33\% & \textbf{85.93\%} & 84.51\%\\
    \cline{2-6}
     & Ours & 82.79\% & 78.98\% & \textbf{86.66\%} & 79.48\% \\
    \hline
     & Sha et al.$_{EtE}$ & 81.54\% & 79.42\% & \textbf{82.24\%} & 80.81\% \\
    \cline{2-6}
    Intersection N.4 & Sha et al. & 80.65\% & 78.53\% & \textbf{83.16\%} & 81.23\% \\
    \cline{2-6}
     & Ours$_{EtE}$ & 83.61\% & 82.64\% & \textbf{83.72\%} & 82.52\%\\
    \cline{2-6}
    & Ours & 82.22\% & 78.98\% & \textbf{84.51\%} & 82.23\% \\
    \hline
     & Sha et al.$_{EtE}$ & 67.82\% & 59.93\% & \textbf{71.91\%} & 63.63\% \\
    \cline{2-6}
    Intersection N.5 & Sha et al. & 67.71\% & 60.43\% & \textbf{71.95\%} & 64.96\% \\
    \cline{2-6}
     & Ours$_{EtE}$ & 69.71\% & 60.55\% & \textbf{73.13\%} & 63.71\%\\
    \cline{2-6}
     & Ours & 71.11\% & 60.03\% & \textbf{73.75\%} & 62.76\% \\
    \hline
\end{tabular}
\label{tab:largemodelscompare}
\end{table*} 

\subsection{Homography estimation with Topological Loss}
To evaluate the performance of Topological Loss, we compare the results obtained by the proposed model and the model proposed by Sha et al. \cite{9157432} (further on called "baseline") in Table~\ref{tab:largemodelscompare}. Applied to the baseline model, Topological Loss causes improvements of 2\% on the average. Such improvements are especially prominent for the intersection datasets, which have more complex topologies. This result is due to the fact that our loss function allows the network to learn the global topology of the scene in a more effective way than a standard pixel-based loss function. The effect of Topological Loss on the proposed model is very similar. For the first intersection fataset, Topological Loss improves MSE and Dice accuracy by 2\%. Such improvement is even more evident for the third and fifth intersections, where Topological Loss improves MSE and Dice by 7\% and 10\%. The performances of topological loss on the fifth intersection is especially relevant due to the complexity of the topology of the scene. 

\begin{table*}[t]
\caption{IoU scores for the mixed dictionary datasets.}
\centering
\begin{tabular}{|l|l|c|c|c|c|}
    \hline
    {\textbf{Dataset}} & {\textbf{Method}} & \multicolumn{4}{c}{\textbf{Measured IoU}}\\
    \cline{3-6}
    & & $\mathcal{L}_{\text{Top-MSE}}$ & MSE & $\mathcal{L}_{\text{Top-Dice}}$ & Dice\\
    \hline
     & Sha et al.$_{EtE}$ & \textbf{72.25\%} & 72.06\% & 72.03\% & 72.19\% \\
    \cline{2-6}
    First Mix & Sha et al. & 72.14\% & 72.12\% & \textbf{72.33\%} & 71.93\% \\
    \cline{2-6}
     & Ours$_{EtE}$ & 86.37\% & 85.48\% & \textbf{87.42\%} & 85.93\%\\
    \cline{2-6}
     & Ours & 85.17\% & 76.87\% & \textbf{86.07\%} & 84.71\% \\
    \hline
     & Sha et al.$_{EtE}$ & 80.48\% & 73.16\% & \textbf{82.05\%} & 78.14\% \\
    \cline{2-6}
    Second Mix & Sha et al. & 80.59\% & 78.47\% & \textbf{81.75\%} & 77.93\% \\
    \cline{2-6}
     & Ours$_{EtE}$ & 83.44\% & 82.71\% & \textbf{84.05\%} & 82.53\%\\
    \cline{2-6}
     & Ours & 84.36\% & 82.84\% & \textbf{84.85\%} & 82.53\%\\
    \hline
     & Sha et al.$_{EtE}$ & 66.65\% & 61.73\% & \textbf{68.01\%} & 59.09\% \\
    \cline{2-6}
    Third Mix & Sha et al. & 68.22\% & 59.92\% & \textbf{69.41\%} & 62.26\% \\
    \cline{2-6}
     & Ours$_{EtE}$ & \textbf{76.64\%} & 60.48\% & 73.15\% & 61.72\%\\
    \cline{2-6}
     & Ours & \textbf{77.82\%} & 60.85\% & 73.34\% & 65.92\% \\
    \hline
\end{tabular}
\label{tab:mixedCompare}
\end{table*}

\subsection{Ablation Study of the Proposed Model}
The comparison between the proposed model and the baseline model shows that the improvements introduced by the location blocks and the self-attention to the STN are valuable, as shown in Table~\ref{tab:largemodelscompare}. The benefits are especially evident in the experiments on the second and third intersection datasets, where the proposed model outperforms the baseline when trained with the topological losses by a margin of almost 10\%. In the experiments, we find that the end-to-end training of the baseline model on the intersection datasets degrades slightly the performance of the model, contrary to what the authors report in their paper. Concerning the World Cup 2014 dataset, the proposed model outperforms the baseline by 3\% for every combination of loss and matching functions. 

\subsection{Mixed dictionary}
The difference in performance between the two models on complex topologies, such as the intersections, suggests that the benefits brought by Localization Blocks and self-attention are very useful to the STN. The proposed model manages to be less dependent on an optimal template match and more capable to learn the topologies from the dataset. To test the capabilities of this model, we experiment with a mixed dictionary. In this setting, the training and testing splits are the same as the one used in previous experiments, while the dictionary is composed of templates randomly sampled from the other intersection's dataset. In this way, the matched template provided to the STN with the segmented map is of a completely different intersection. We construct three mixed dictionaries to use with the training and testing splits of the first, fourth and fifth intersections by sampling images from the other four intersections.
The results in Table~\ref{tab:mixedCompare} reveal that the proposed model handles these situations well, improving on the baseline model performance by up to~17\%. \\
\indent
Comparing Tables~\ref{tab:largemodelscompare} and \ref{tab:mixedCompare}, the proposed model obtains better result on the third intersection when trained the mixed dictionary approach. A similar behaviour can be observed by comparing Tables~\ref{tab:largemodelscompare} and \ref{tab:mixedCompare}. In this case, the proposed model performs slightly better in the mixed dictionary setting while the baseline performances decrease. This is especially evident when comparing the performances of the models trained with Dice loss and MSE. In fact, Tables~\ref{tab:largemodelscompare} and \ref{tab:mixedCompare} show that the improvements brought by Topological loss are crucial to the performances of the models. Both models perform definitely worse when trained with Dice loss and MSE. This is probably due to the fact that these loss functions are unequipped to deal with the complexity of the topology of the fifth intersection when there are no optimal templates available. We conjecture that the proposed model has proper generalization capabilities across datasets, but the dictionary-based matching process prevents this approach to scale. The Topological loss improves the performances of the baseline model as well, as shown in Table~\ref{tab:largemodelscompare}.

\section{Conclusion}
This research proposes an improved block structure (LocBlock) of the standard localization layers part of an STN model and a new loss function for performing a homography estimation to establish automated camera calibration. 
The overall architecture includes a UNet to perform semantic segmentation on the input images, a Siamese network to match the semantic maps with a dictionary of semantic templates with known homographies, and a modified STN, to estimate a homography matrix. All experiments show that the combination of the proposed model and the proposed topological loss function improves upon previous the state-of-the-art approach on every dataset we tested on. We show that the combination of the proposed model and the topological loss function is also flexible to the composition of the dictionary, outperforming every other combination of model and loss function. \\
\indent
The synthetic dictionary approach allows the model to be trained on synthetic homographies. Therefore, the model does not require the ground-truth homographies to be trained on real-world datasets, only the semantic ground-truth images. This is especially relevant given that gathering and using images and videos of traffic can result in ethical and legal issues. 
The loss function leverages the pixel-based errors between patches of the warped image and the ground-truth patches by increasing the error of that patch with the errors of the neighbouring patches. \\
\indent
A major problem of the experiments is the lack of available real-world data sets on which the proposed model and loss functions can be tested. In addition, the Siamese network is a critical component for the architecture due to the dictionary-based approach, which is computationally expensive and does not easily scale-up for large datasets such as the complete intersections dataset.
Additional work in this direction should focus on a more efficient way to perform the matching process.

\bibliographystyle{ieeetr}
\bibliography{egbib}

\end{document}